\documentclass[letterpaper]{article} % DO NOT CHANGE THIS
\usepackage[]{aaai24}  % DO NOT CHANGE THIS
\usepackage{times}  % DO NOT CHANGE THIS
\usepackage{helvet}  % DO NOT CHANGE THIS
\usepackage{courier}  % DO NOT CHANGE THIS
\usepackage[hyphens]{url}  % DO NOT CHANGE THIS
\usepackage{graphicx} % DO NOT CHANGE THIS
\usepackage{natbib}  % DO NOT CHANGE THIS AND DO NOT ADD ANY OPTIONS TO IT
\usepackage{caption} % DO NOT CHANGE THIS AND DO NOT ADD ANY OPTIONS TO IT
\usepackage{subfiles}
\usepackage{multirow}
\usepackage{amsmath}
\usepackage{subcaption}
\usepackage{mathtools}
\usepackage{booktabs}
\usepackage{amsthm}
\usepackage{amsfonts}
\usepackage{gensymb}
\usepackage{tikz}
\usepackage{algorithm}
\usepackage{algorithmic}

\usepackage{newfloat}
\usepackage{listings}
\DeclareCaptionStyle{ruled}{labelfont=normalfont,labelsep=colon,strut=off} % DO NOT CHANGE THIS
\floatstyle{ruled}
\newfloat{listing}{tb}{lst}{}
\floatname{listing}{Listing}
\newcommand{\xm}{%
\tikz[scale=0.23] {
    \draw[line width=0.7,line cap=round] (0,0) to [bend left=6] (1,1);
    \draw[line width=0.7,line cap=round] (0.2,0.95) to [bend right=3] (0.8,0.05);
}}
\newcommand{\cm}{%
\tikz[scale=0.23] {
    \draw[line width=0.7,line cap=round] (0.25,0) to [bend left=10] (1,1);
    \draw[line width=0.8,line cap=round] (0,0.35) to [bend right=1] (0.23,0);
}}
\title{Enhancing Motion Variation in Text-to-Motion Models via Pose and Video Conditioned Editing}

\author{
    Clayton Souza Leite, Yu Xiao
}
\affiliations{
    Aalto University\\
    \{clayton.souzaleite\}\{yu.xiao\}@aalto.fi
}

\usepackage{bibentry}
\begin{document}

\maketitle

\begin{abstract}

Text-to-motion models that generate sequences of human poses from textual descriptions are garnering significant attention. However, due to data scarcity, the range of motions these models can produce is still limited.
%Text-to-motion models synthesize sequences of human poses from textual descriptions but are severely constrained by the limited availability of datasets. This data scarcity hinders the models' natural language understanding, restricting the range of motions they can generate. 
For instance, current text-to-motion models cannot generate a motion of kicking a football with the instep of the foot, since the training data only includes martial arts kicks. We propose a novel method that uses short video clips or images as conditions to modify existing basic motions. In this approach, the model's understanding of a kick serves as the prior, while the video or image of a football kick acts as the posterior, enabling the generation of the desired motion. By incorporating these additional modalities as conditions, our method can create motions not present in the training set, overcoming the limitations of text-motion datasets. A user study with 26 participants demonstrated that our approach produces unseen motions with realism comparable to commonly represented motions in text-motion datasets (e.g., HumanML3D), such as walking, running, squatting, and kicking.
\end{abstract}

\section{Introduction}

\iffalse
\begin{figure*}
    \centering
    \includegraphics[width=0.98\textwidth]{figures/cover_all.png}
    \caption{Our method generates new motions from a base motion and an inserted pose or short video clip. For instance, given a base motion of walking lightly (depicted on the left), the user can adjust the body joints via a GUI to insert a pose of barely holding a heavy load. The method then transforms the motion to depict short, burdened steps. Alternatively, a short video clip (2-4 seconds) can be used. Our method can modify the common kick into a football kick using information from a low-quality video of a person kicking a football, despite the absence of such types of kicks in the training dataset (HumanML3D \cite{humanml3d}). The final motion mimics the motion of kicking a football with the instep of the foot like in the video input.}
    \label{fig:cover}
\end{figure*}
\fi

\begin{figure*}[t]
    \centering
    \begin{subfigure}[b]{0.49\textwidth}
        \centering
        \includegraphics[width=\textwidth]{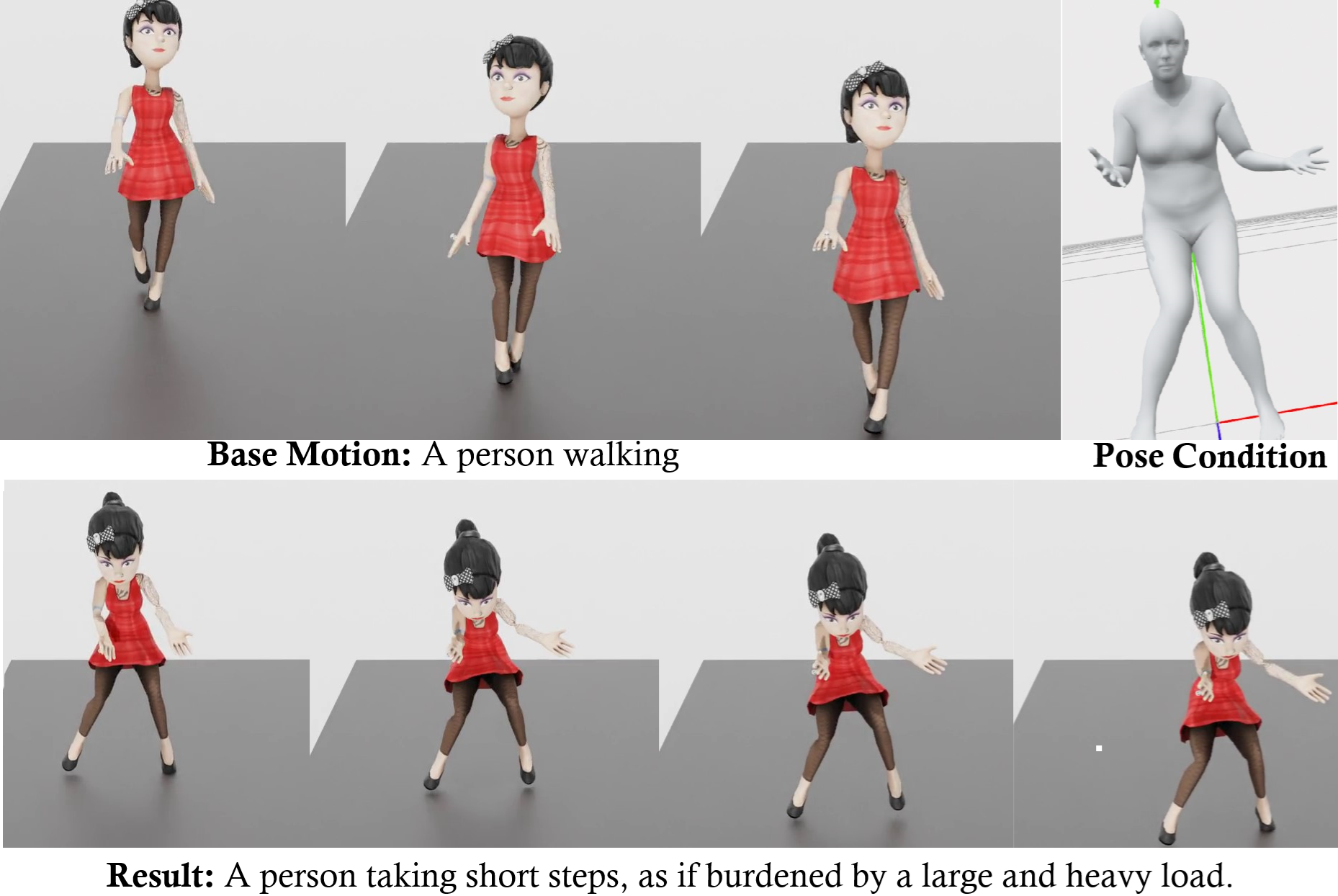}
        \caption{Pose Insertion}
        \label{fig:posecond}
    \end{subfigure}
    \hfill
    \begin{subfigure}[b]{0.49\textwidth}
        \centering
        \includegraphics[width=\textwidth]{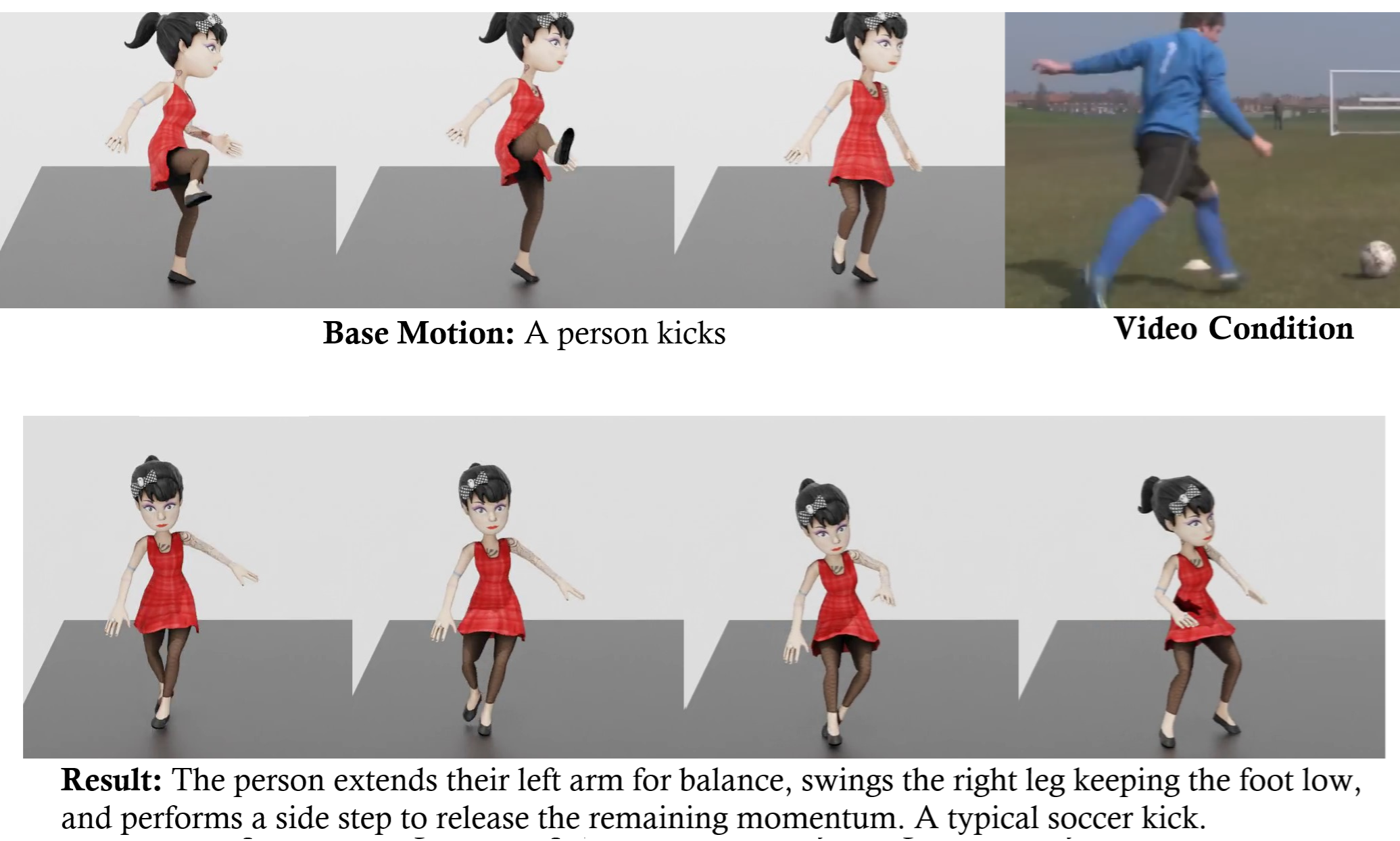}
        \caption{Video Insertion}
        \label{fig:videocond}
    \end{subfigure}
    \caption{Our method generates new motions from a base motion and an inserted pose or short video clip. For instance, given a base motion of walking lightly (depicted on the left), the user can adjust the body joint angles via a GUI (developed by \citet{bodymodelgithub}) to insert a pose of barely holding a heavy load. The method then transforms the motion to depict short, burdened steps. Alternatively, a short video clip (2-4 seconds) can be used. Our method can modify the common kick into a football kick using information from a low-quality video of a person kicking a football, despite the absence of such types of kicks in the training dataset (HumanML3D \cite{humanml3d}). The final motion mimics the motion of kicking a football with the instep of the foot like in the video input.}
    \label{fig:cover}
\end{figure*}

\iffalse
\begin{figure*}[t]
    \centering
    \begin{subfigure}[b]{0.49\textwidth}
        \centering
        \includegraphics[width=\textwidth]{figures/covera.png}
        \caption{Pose Insertion}
        \label{fig:posecond}
    \end{subfigure}
    \hfill
    \begin{subfigure}[b]{0.49\textwidth}
        \centering
        \includegraphics[width=\textwidth]{figures/coverb.png}
        \caption{Video Insertion}
        \label{fig:videocond}
    \end{subfigure}
    \caption{Our method generates new motions from a base motion and an inserted pose or short video clip. For instance, given a base motion of walking lightly (depicted on the left), the user can adjust the body joint angles via a GUI (developed by \citet{bodymodelgithub}) to insert a pose of barely holding a heavy load. The method then transforms the motion to depict short, burdened steps. Alternatively, a short video clip (2-4 seconds) can be used. Our method can modify the common kick into a flying sidekick using information from a low-quality video of a person performing such a kick, despite the absence of such types of kicks in the training dataset (HumanML3D \cite{humanml3d}). The final motion mimics the motion of jumping and performing a side kick in the air.}
    \label{fig:cover}
\end{figure*}
\fi

Recently, there has been a significant surge of interest in generative AI models that enable the creation of diverse content such as images, videos, and human motion directly from textual prompts. %Human motion generation involves synthesizing a sequence of human poses guided by specific conditions. This approach has also seen a notable increase in popularity with textual descriptions as the primary condition for human motion generation. 
Models that translate text into motion (i.e., a sequence of human poses) \cite{mdm} are known as text-to-motion models and are trained on a dataset containing pairs of textual descriptions and corresponding pose sequences. 

Text-to-motion models are trained using datasets that pair specific motions with their corresponding textual descriptions. However, collecting high-quality motion data presents significant challenges. While motion capture systems can provide precise motion data, they are expensive and complex to use. On the other hand, the latest computer vision techniques (e.g., \citet{mediapipe, pare}) for pose estimation using a single camera still yield noisy estimations, necessitating additional manual effort. Consequently, text-to-motion models face significant data scarcity. This scarcity hinders the models' performance, making it challenging for them to capture and reproduce the intricacies and variations of human motion described in the input text (e.g., various types of kicks or different forms of walking).

Certain text-to-motion models \cite{mdm, momask, liu2023plan, flame} enable the modification of an already generated motion by conditioning specific body joints or temporal portions of the motion on a different text prompt. This feature helps introduce variation into a given motion, partially compensating for the aforementioned limitation of text-to-motion models. For instance, to generate a motion of walking while carrying a heavy bag with two hands, one could first generate a walking motion and then modify it by conditioning upper-body joints using the text prompt ``a person is carrying a heavy bag with two hands." 
%one can generate a walking motion and then edit it by conditioning the upper-body joints with the text prompt ``a person is holding something." This results in a motion where the person appears to be holding an object while walking. 
We call this \textbf{local motion edition}, where only local (spatial or temporal) characteristics of the generated motion are modified. The extent to which new motions can be generated through local motion edition is limited because models, due to data scarcity, have a small vocabulary of motions and do not fully comprehend variations in motions. For example, the models cannot interpret how a person would carry the bag and are unable to adjust the poses to accommodate variations in the size, shape, or weight of the bag relative to the person. %Existing models do not support \textbf{global motion edition}. 
%However, text-conditioned local motion editing still results in limited generation since these models, as a consequence of scarce datasets, have a relatively small vocabulary of motions and do not fully comprehend variations in motions. For example, it is not possible to specify how the person is holding the load (over the head or in front of the person) or the size of the load (whether it requires the person to open their arms widely or is small enough to be carried in their hands). 
%Moreover, these models do not support the edition of global characteristics of a generated motion. For instance, carrying a heavy load affects all joint angles, resulting in an atypical walking style that current text-to-motion models cannot generate due to their limited understanding of human motion variations.

To enhance the variations of the generated motions, we propose a novel method that enables the editing of local and global characteristics of an existing motion using a pose or video as a condition, unlike previous works that use text as a condition.
 %To the best of our knowledge, our approach is the first to create motions by editing \textbf{local or global} characteristics of an existing motion using a pose or video condition, rather than a text condition. 
For example, to generate the motion of a person taking short steps as if burdened by a heavy load, our method uses as input the motion of a person walking and the human pose of someone carrying a large object with bent knees, indicating the weight. The method then generates a walking style in which the input pose is consistently integrated (Fig. \ref{fig:posecond}). We refer to this as \textbf{global motion edition} because it involves altering the entire body’s behavior throughout the action of walking -- i.e., from the hands holding the imaginary object to the knees bending to indicate a heavy load. To the best of our knowledge, this is the first work that supports global motion edition.

 In the case of a video condition, such as a person kicking a football, our method takes a single short (2-5 seconds) video clip of a person kicking a football as input along with the base motion of kicking. This results in an output that accurately reflects the football kick (Fig. \ref{fig:videocond}). Utilizing pose or video as conditions for motion editing not only provides better control over the final desired motion -- due to the richer details present in images and videos compared to text -- but also circumvents the limitations of current text-to-motion models in understanding prompts due to the limited dataset they were trained with. 
 %In the case of a video condition, such as a person performing a flying sidekick, our method takes a short (2-5 seconds) video clip of such an action as input along with the base motion of kicking. This results in an output that reflects the kick in the video (Fig. \ref{fig:videocond}). Utilizing pose or video as conditions for motion editing not only provides better control over the final desired motion --due to the richer details present in images and videos compared to text -- but also circumvents the limitations of current text-to-motion models in understanding prompts due to the limited dataset they were trained with.

\iffalse
\begin{figure*}
    \centering
    \includegraphics[width=0.9\textwidth]{figures/cover2.png}
    \caption{Our method also generates motions from a base motion and a short video clip (2-4 seconds). Although the base motion depicts a person kicking, it does not resemble a typical football kick. Moreover, the model is unable to generate this type of kick since the training dataset (HumanML3D \cite{humanml3d}) does not include it. By inserting a low-quality video from the YouTube-8M dataset, our method utilizes the video's information to modify the base motion into a football kick.}
    \label{fig:cover2}
\end{figure*}
\fi

To evaluate the realism of the motions generated by our motion editing methods, we conducted a quantitative user study with 26 participants. These participants were presented with 16 motions generated by our approach (listed in Table \ref{tab:pi} and \ref{tab:vi}), as well as basic motions produced by a state-of-the-art diffusion-based text-to-motion model \cite{mdm}, such as walking, running, sitting down, kicking, and squatting. These basic motions, commonly represented in text-motion datasets, are known to be generated with high realism by existing models. Participants were asked to rate the realism of the motions. The study revealed that our method produced novel motion variations with high realism, comparable to the basic motions, without requiring extensive amounts of data.

The rest of the paper is organized as follows. Section 2 covers the related work and background. Section 3 introduces our method. Section 4 presents the experiments. Section 5 discusses the benefits, limitations, and future work. Finally, Section 6 concludes the work.

\section{Related Work and Background}

\iffalse
\begin{table*}[]
\centering
\begin{tabular}{lcc}
\hline
\textbf{Methods} &
  \textbf{Local Motion Edition} &
  \textbf{Global Motion Edition} \\ \hline
\begin{tabular}[c]{@{}l@{}}TEMOS, MotionDiffuse, FG-T2M, T2M-GPT, \\ MotionGPT, AttT2M, HumanTOMATO\end{tabular} &
  \xm & \xm
   \\ \hline
MDM, Pro-Motion, FLAME, MoMask & \cm & \xm \\ \hline
Ours                           & \cm & \cm \\ \hline
\end{tabular}
\caption{Comparison of our method with the latest text-to-motion techniques regarding motion editing capabilities.}
\label{tab:compare}
\end{table*}
\fi

%In this section, we review the related work in text-to-motion models -- including their motion edition capabilities -- and provide the necessary background to understand the concepts of our method. 

This section reviews  two categories of text-to-motion models -- discrete latent space and diffusion-based -- and their motion editing capabilities. In addition, it covers the fundamentals of diffusion-based methods employed in this work.

\subsection{Discrete Latent Space Methods}
Discrete latent space methods represent motion as sequences of discrete codes and map text descriptions to these sequences. 
T2M-GPT \cite{t2mgpt} proposed a two-stage motion generation from text using VQ-VAE (Vector Quantised-Variational Auto Encoder) \cite{vqvae} for mapping motion to discrete codes and GPT \cite{gpt} for generating code sequences from text. %The method achieved performance comparable to or better than MDM and MotionDiffuse on HumanML3D and KIT-ML datasets.
%VQ-VAE (Vector Quantised-Variational Auto Encoder) \cite{vqvae} primarily distinguishes itself from Variational Auto Encoders (VAE) by proposing an encoder that generates a discrete latent space, as opposed to a continuous one, and by learning the prior distribution instead of assuming a Gaussian distribution. 

MotionGPT \cite{motiongpt} integrated motion (in the form of VQ-VAE quantized tokens) and text into a unified vocabulary, leveraging pre-trained large language models fine-tuned on this combined vocabulary for improved performance over T2M-GPT and MDM. AttT2M \cite{attt2m} also focused on language modeling, proposing an improved architecture with sentence-level self-attention and word-level cross-attention. MoMask \cite{momask} proposed to enhance the VQ-VAE quantization by reducing approximation errors through residual vector quantization. 

%If HumanTOMATO does not support edition, we can remove it, since we already have many examples in this part.

%HumanTOMATO \cite{humantomato} generated whole-body motion, including hands and facial expressions, using two-level discrete codes for body and hand, and a continuous latent space for facial expressions.

%Previous works have concentrated on generating motion from text. However, since they rely on the same datasets (e.g., HumanML3D), the range of motions they can produce is consistently limited across all methods due to the scarcity of text-motion data. 

Although MoMask supports the editing of generated motions, its editing capabilities are restricted to textual conditions and only affect local temporal characteristics (i.e., only a specific temporal segment can be modified). 
%We emphasize that motion editing conditioned on text cannot expand the range of motions due to the limited language-to-motion understanding of current text-to-motion models. 
In contrast, our method allows for both global and local motion editing, conditioned on images, videos, or a pose provided in a graphical user interface.

\subsection{Diffusion-Based Methods}

Diffusion-based methods use diffusion models \cite{guideddiffusion} to generate motion from text by starting with random noise and iteratively denoising it guided by a text prompt. MotionDiffuse \cite{motiondiffuse}, the first diffusion-based framework for motion synthesis from text, used Denoising Diffusion Probabilistic Models (DDPM). The method proposed to decompose the whole-body motion into semi-independent parts (e.g., the upper and lower body), enabling better control over each part for motion synthesis.

MDM \cite{mdm} devised a lightweight diffusion model with additional loss terms to enforce realistic physical properties. FG-T2M \cite{fgt2m} focused on language modeling in motion synthesis, incorporating modules for word and syntax analysis for better context understanding. PRO-Motion \cite{liu2023plan} used GPT-3.5 \cite{gpt} to convert text prompts into key pose descriptions. A diffusion-based model generated poses from these descriptions, which were then interpolated into coherent motion by another diffusion-based model. 

%Similarly, the diffusion-based methods discussed focus on generating realistic motion from textual prompts. 

While MDM, PRO-Motion, and FLAME offer editing capabilities on the generated motions, their editing is restricted to local temporal or spatial (e.g., specific joints)  characteristics based on text prompts. Altering global aspects of a motion, such as the overall style of walking or kicking a ball, is not feasible with these methods. Our approach is not restricted to local motion editing and supports editing based on video or pose conditions, rather than text. By using video or pose information as a condition, our method can generate motions not present in the training set. %Our method leverages the latest advances in diffusion-based techniques, which have demonstrated impressive editing capabilities conditioned on visual input \cite{guideddiffusion, imagic, unipaint}. To provide context, we first present the mathematical background on these methods.

\subsection{Fundamentals of Diffusion-Based Methods}

Our method leverages the latest advances in diffusion-based techniques, which have demonstrated impressive editing capabilities conditioned on visual input \cite{guideddiffusion, imagic, unipaint}. To provide context, we first present the mathematical background on these methods.

We denote $\pmb{x}_0$ as a sequence of human poses (i.e., human motion), where each pose is represented by XYZ body joint locations, rotations of the body joints, or a combination of both. Additionally, the pose may include velocities in spatial or joint space. Hence, $\pmb{x}_0 \in \mathbb{R}^{N \times D}$ where $N$ is the number of time steps in the pose sequence and $D$ is the dimension of the pose representation. In our approach, we use the motion representation by HumanML3D \cite{humanml3d}, which redundantly includes XYZ joint locations, XYZ joint velocities, and joint rotations; resulting in $D = 263$. HumanML3D uses 23 body joints. 

%In diffusion models, a synthetic motion is initialized as randomly sampled noise, denoted as $\pmb{x}_T \sim \mathcal{N}(0, \pmb{I})$, and is subsequently denoised through iterative steps $\pmb{x}_t$ for $t \in \{T, T - 1, ..., 0\}$, ultimately achieving a realistic motion $\pmb{x}_0$ \cite{imagic}.  
The diffusion process is modeled as a Markov chain that gradually adds noise to the data according to a variance schedule given by $0 < \beta_t < 1$, resulting in $\pmb{x}_t$, which is a noised version of $\pmb{x}_0$.

\begin{equation}
q(\pmb{x}_t|\pmb{x}_{t-1}) := \mathcal{N}(\pmb{x}_{t}, \sqrt{1 - \beta_t}  \pmb{x}_{t-1},\beta_t \pmb{I})
\end{equation}

$\pmb{I}$ denotes the identity matrix. Instead of adding noise iteratively from $\pmb{x}_0$ to $\pmb{x}_t$, \citet{ddpm} provides the closed form given in Eq. \ref{eq:closed}, where $\bar{\alpha_t} = \prod_s^{t} 1 - \beta_s$ and $\pmb{\epsilon} \sim  \mathcal{N}(\pmb{0}, \pmb{I})$.
 
\begin{equation}
q(\pmb{x}_t | \pmb{x}_0) = \sqrt{ \bar{\alpha_t}}\pmb{x}_0 +  \pmb{\epsilon} \sqrt{1 - \bar{\alpha_t}}
\label{eq:closed}
\end{equation}

The reverse diffusion process is also modeled as a Markov chain (Eq. \ref{eq:p_}) where the $\pmb{x}_t$ is progressively refined through noise removal. Each refinement step (i.e., to obtain $\pmb{x}_{t-1}$ from $\pmb{x}_t$) consists of applying a neural network $f_{\theta}(\pmb{x}_t, t, \pmb{c})$, where $\pmb{c}$ is a condition (e.g., textual description), to predict the mean $\pmb{\mu}_{\theta}(\pmb{x}_t, t)$ and variance $\pmb{\Sigma}_{\theta}(\pmb{x}_t, t)$. A common approach is to fix the variance to $\pmb{\Sigma}_{\theta}(\pmb{x}_t, t) = \frac{1 - \bar{\alpha}_{t-1}}{1 - \bar{\alpha_t}}\beta_t$ \cite{ddpm}.

\begin{equation}
p_{\theta}(\pmb{x}_{t-1}|\pmb{x}_t) := \mathcal{N}(\pmb{x}_{t-1}, \pmb{\mu}_{\theta}(\pmb{x}_t, t), \pmb{\Sigma}_{\theta}(\pmb{x}_t, t))
\label{eq:p_}
\end{equation}

The loss function in Eq. \ref{eq:loss} is used to train the neural network $f_{\theta}$ to predict the mean $\pmb{\mu}_{\theta}(\pmb{x}_t, t)$. The ground-truth mean is given by $\pmb{\tilde{\mu}}(\pmb{x}_t, \pmb{x}_0) = \frac{\sqrt{\bar{\alpha}_{t - 1}}\beta_t}{1 - \bar{\alpha}_t} \pmb{x}_0 + \frac{\sqrt{\alpha_t} (1 - \bar{\alpha}_{t-1})}{1 - \bar{\alpha}_t} \pmb{x}_t$.

\begin{equation}
L = || \pmb{\tilde{\mu}}(\pmb{x}_t, \pmb{x}_0)  - \pmb{\mu}_{\theta}(\pmb{x}_t, t) ||^2 
\label{eq:loss}
\end{equation}

An alternative to predicting the mean is to predict $\pmb{x}_0$ instead of $\pmb{\mu}_{\theta}(\pmb{x}_t, t)$. This allows direct operations on the motion within the loss function. For instance, by using $f_{\theta}$  to predict $\pmb{x}_0$, we can mask the loss function so the neural network learns to ignore specific time steps or components in  $\pmb{x}_0$.  This is useful when  $\pmb{x}_0$ contains systematic noise, such as due to the presence of motion obtained through human pose estimation methods -- i.e., our case. The loss function to predict $\pmb{x}_0$ is given in Eq. \ref{eq:loss2}. %This concept is further clarified in the \textbf{Embedding Optimization Module}. 

\begin{equation}
L = || \pmb{x}_0  - f_{\theta}(\pmb{x}_t, t, \pmb{c})||^2 
\label{eq:loss2}
\end{equation}

Note that predicting $\pmb{x}_0$ from $\pmb{x}_t$ is generally unreliable (except when $t = 1$) due to the high level of noise present in $\pmb{x}_t$. Therefore, it remains necessary to iteratively denoise $\pmb{x}_t$ by calculating the mean from the estimated value of $\pmb{x}_0$ and $\pmb{x}_t$ as in Eq. \ref{eq:meanxstart}.

\begin{equation}
\pmb{\mu}(\pmb{x}_t, \pmb{x}_0) = \frac{\sqrt{\bar{\alpha}_{t - 1}}\beta_t}{1 - \bar{\alpha}_t} f_{\theta}(\pmb{x}_t, t, \pmb{c}) + \frac{\sqrt{\alpha_t} (1 - \bar{\alpha}_{t-1})}{1 - \bar{\alpha}_t} \pmb{x}_t
\label{eq:meanxstart}
\end{equation}

The calculation of $\pmb{\mu}(\pmb{x}_t, \pmb{x}_0)$ via Eq. \ref{eq:meanxstart} allows us to obtain $\pmb{x}_{t-1}$ by applying Eq. \ref{eq:p_}.

\section{Methodology}

In this section, we provide an overview of our method, a detailed explanation of each of its components, and additional implementation details.

%\subsection{Problem Formulation}
\subsection{System Overview}

\begin{figure}[t]
    \centering
    \includegraphics[width=0.45\textwidth]{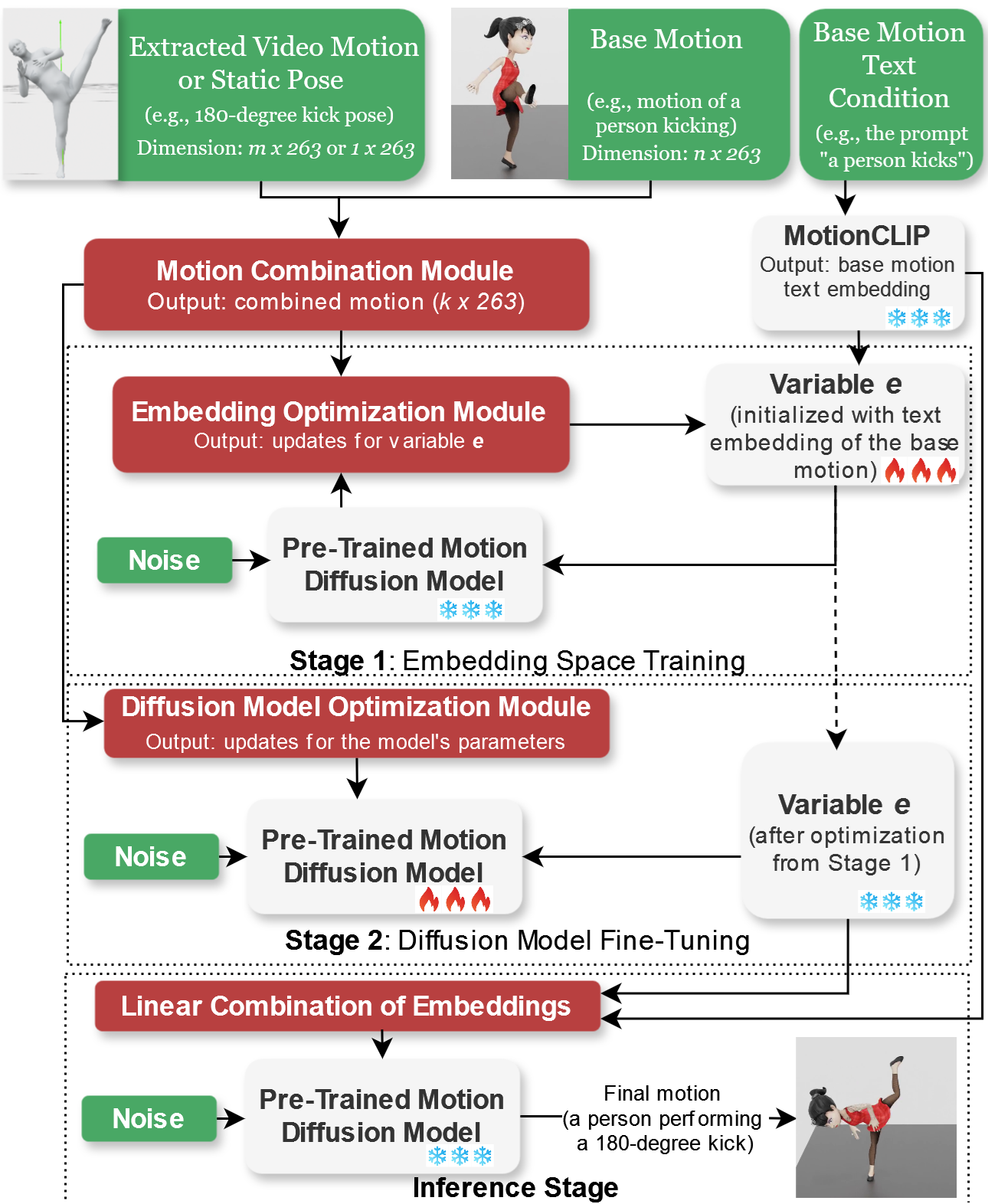}
    \caption{Overview of our method. Green blocks represent data, red blocks process the data, and gray blocks are neural network models or parameters. Ice and fire icons indicate frozen and trainable parameters, respectively.}
    \label{fig:method}
\end{figure}

As illustrated in Fig. \ref{fig:method}, the system takes three inputs:  a video clip or a static pose, a base motion $\pmb{x}_0^{B}$, and the text condition corresponding to the base motion. The base motion is then dynamically transformed to either match certain features or incorporate new attributes dictated by the condition, resulting in a new, modified motion sequence that blends the original movement with the desired changes. This process is implemented with two training stages and an inference stage, inspired by Imagic \cite{imagic}. The first training stage focuses on training the embedding space, whereas the second fine-tunes the diffusion model.
Our system supports both local and global modifications. Unlike local modifications, which are confined to a particular temporal segment of $\pmb{x}_0^{B}$, global modifications require $\pmb{x}_0^{B}$ to adapt its overall characteristics to align with the input condition.

%For local modifications, a specific temporal segment of $\pmb{x}_0^{B}$ must be adjusted to align with the input condition. In contrast, global modifications require $\pmb{x}_0^{B}$ to integrate the overall characteristics of the input condition.

%The base motion is the motion that will be modified according to the video or static pose conditions.

%Given a base motion $\pmb{x}_0^{B}$ and an image or video condition, the goal is to modify $\pmb{x}_0^{B}$ by altering its local or global characteristics.  

%Fig. \ref{fig:method} illustrates our pipeline. The input consists of three components: a video or a static pose, a base motion, and the text condition corresponding to the base motion. The base motion is the motion that will be modified according to the video or static pose conditions. 

\subsubsection{Preprocessing of Input}
Given a video clip as the condition, SMPL \cite{smpl} (Skinned Multi-Person Linear model) parameters are first extracted from each video frame using PARE \cite{pare}. These parameters, which consist of XYZ coordinates for 6890 vertices, are then converted into the motion representation, specifically the HumanML3D motion representation in our case.

If the input is an image capturing a static pose, the pose
can be extracted by processing the image with PARE \cite{pare} and converting the resulting SMPL parameters into the HumanML3D motion representation. Alternatively, the static pose can be manually adjusted using software that allows users to modify joint angles. We use the Body Model Visualizer software \cite{bodymodelgithub} for this purpose in this work.

%The pipeline consists of two training stages and an inference stage, inspired by Imagic \cite{imagic}. The first training stage focuses on training the embedding space, whereas the second fine-tunes the diffusion model.

\subsubsection{Stage 1: Embedding Space Training}

 In the first stage -- i.e., Embedding Space Training -- the motion extracted from the video or the static pose -- denoted as  $\pmb{x}_0^{I}$ -- is combined with the base motion $\pmb{x}_0^{B}$ in the \textbf{Motion Combination Module} to form a single motion $\pmb{x}_0^{C}$ referred to as the combined motion. Simultaneously, the text condition of the base motion is processed through MotionCLIP \cite{motionclip} to obtain a text embedding, which is used to initialize the trainable variable $\pmb{e}$. Using the \textbf{Embedding Optimization Module}, $\pmb{e}$  is trained to ensure that, when provided as input to a pre-trained motion diffusion model (kept frozen during the first stage), it produces a motion resembling the combined motion $\pmb{x}_0^{C}$. We denote the optimized variable $\pmb{e}$ as $\pmb{e}_{opt}$. Initializing $\pmb{e}$ with the text embedding of the base motion ensures that $\pmb{e}_{opt}$ lies in the vicinity of this initial embedding. This approach preserves the characteristics of the base motion while incorporating those of the combined motion.

\subsubsection{Stage 2: Diffusion Model Fine-Tuning}

Similar to the first stage, the second stage uses the combined motion $\pmb{x}_0^{C}$ and the trained embedding $\pmb{e}_{opt}$ (which is kept frozen this time). In this stage, the neural network representing the diffusion model is fine-tuned with the help of the \textbf{Diffusion Model Optimization Module}. The objective remains the same as in the first stage: when the motion diffusion model receives $\pmb{e}_{opt}$ as input, it must produce a motion resembling the combined motion. The second stage is necessary because training solely the embedding space is insufficient for achieving a motion that resembles the combined motion. This is particularly true if the combined motion includes pose sequences that were not present in the training set -- i.e., the diffusion model has not been trained to generate.

\subsubsection{Stage 3: Inference}

The optimized embedding from Stage 1, the fine-tuned diffusion model from Stage 2, and the embedding of the base motion $\pmb{e}_{base}$ are used to obtain the final desired motion. The base motion and the optimized embeddings are linearly combined according to Eq. \ref{eq:lce} before being passed to the diffusion model to generate the final motion.

\begin{equation}
    \pmb{\bar{e}} \leftarrow \pmb{e}_{opt} \cdot \eta + \pmb{e}_{base} \cdot (1 - \eta)
    \label{eq:lce}
\end{equation}

where $\eta \in [0, 1]$. For lower values of $\eta$, the generated motion closely resembles the base motion. As $\eta$ increases, different levels of the combined motion are integrated into the base motion. However, this also means that the noise present in the combined motion is likely to be carried over into the generated motion. Determining the optimal value for $\eta$, which will produce the best-generated motion, is challenging due to the black-box nature of the model and heavily depends on the hyper-parameters set during the two training stages. In practice, after the second stage, one can generate various motions with different $\eta$ values and select the best motion based on visual inspection. Hence, this linear combination allows for quicker additional control over the generated motions without the need for re-training with different sets of hyper-parameters.

\subsection{Module Description}

%Below we describe each of the aforementioned modules.

\subsubsection{Motion Combination Module}

The combination of the base motion $\pmb{x}_0^{B}$ and the input motion $\pmb{x}_0^{I}$ (which, in the case of a static pose, contains only one time step) depends on the nature of the latter (i.e., whether it is extracted from video or a static pose) and the scenario (i.e., global or local). As mentioned in the Introduction, a local scenario involves modifying only a specific part of the motion, whereas a global scenario modifies the entire motion's characteristics. For a static pose as input, $\pmb{x}_0^{C}$ is defined as in Eq. \ref{eq:local}. 

\begin{equation}
    \pmb{x}_0^{C} = (\pmb{1}_{N} \otimes \pmb{x}_0^{I}) \odot  \pmb{M} + \pmb{x}_0^{B} \odot (\pmb{1} - \pmb{M})
    \label{eq:local}
\end{equation}

where $\pmb{M} \in \mathbb{R}^{N \times D}$ is a mask, $\pmb{1}_{N}$ is a vector of size $N$ with ones, $ \otimes $ is the Kronecker product, and $\odot$ is the Hadamard (element-wise) product. The components $M_{ij}$ of $\pmb{M}$ are defined according to Eq. \ref{eq:mask}.

\begin{equation}
    M_{ij} = 
    \begin{cases} 
    1 & \text{if }  i \in \mathcal{P}  \\ 
    0  &  \text{otherwise}  \end{cases}
    \label{eq:mask}
\end{equation}

%where $\mathcal{R}$ is the set of indices within the pose representation that correspond to rotational components. As previously mentioned, $\pmb{x}_0^{C}$ is a sequence of poses, with each pose represented by XYZ components, rotational components, and velocity components. By setting the mask values to one exclusively for the rotational components, we ensure that the combined motion preserves the translational and velocity characteristics of the base motion.  j \in \mathcal{R} \text{ and }

where $\mathcal{P}$ is the set of time steps during which the static pose $\pmb{x}_0^{I}$ should be present. For global scenarios, $\mathcal{P}$ contains all time steps. For local scenarios, $\mathcal{P}$ contains only a single time step defined by the user. For example, consider a base motion where a person is jumping, and the static pose represents a person with legs open sideways. The goal is to create a motion where the person jumps and opens their legs sideways. If, in the base motion, the person is airborne at time step 30, then $\mathcal{P} = \{30\}$, meaning that the static pose will be inserted at this specific time step. Hence, the user makes this choice by observing the base motion and identifying the appropriate time step to insert the static pose.

In the case of video-extracted motion as input, $\pmb{x}_0^{C} = \pmb{x}_0^{I}$ in the global scenario. In the local scenario, $\pmb{x}_0^{C}$ is defined as in Eq. \ref{eq:global}.

\begin{equation}
    \pmb{x}_0^{C} = \begin{bmatrix}
    \mathbf{0}_{P} \\
\pmb{x}_0^{I} \\
\mathbf{0}_{N + P }
\end{bmatrix}  \odot  \pmb{M} + \pmb{x}_0^{B}  \odot (\pmb{1} - \pmb{M})
 \label{eq:global}
\end{equation}

where $P$ is the time step where the video-extracted motion is inserted into the base motion, $N$ is the number of time steps in the base motion, and $\pmb{M}$ is a mask also defined by Eq. \ref{eq:mask} with $\mathcal{P} = \{P, P+1, \ldots, N+P\}$.

\subsubsection{Embedding Optimization Module}

The optimization of the embedding is performed by minimizing the loss function specified in Eq. \ref{eq:loss_emb}
\begin{equation}
    L_{\epsilon} = \pmb{W}  \odot ||\pmb{x}_0^C  - f_\theta(\pmb{x}_t^{C}, t, \pmb{e}) ||^2
    \label{eq:loss_emb}
\end{equation}

where $\pmb{W} \in \mathbb{R}^{N \times D}$ is a weighting term applied to the components of the loss, which varies depending on the nature of the input and the specific scenario. The components $W_{ij}$ in $\pmb{W}$ are defined in Eq. \ref{eq:wloss} for the global scenario or Eq. \ref{eq:wloss2} for the local one. Additionally, $t$ $\sim$ $ U(1, T)$, $\pmb{x}_t^{C}$ is a noisy version of $\pmb{x}_0^{C}$ obtained via Eq. \ref{eq:closed}, and $f$ is the pre-trained diffusion model parameterized by the weights $\theta$, which remain frozen during the optimization of the embedding.

\begin{equation}
W_{ij} = 
\begin{cases} 
1 & \text{if } j \in \mathcal{R}  \\ 
0  &  \text{otherwise}  \end{cases}
\label{eq:wloss}
\end{equation}

\begin{equation}
W_{ij} = 
\begin{cases} 
v  & \text{if }  i = P_m \text{ and }  j \in \mathcal{R}  \\ 
0 & \text{if }  i \in \mathcal{P'} \\ 
1  &  \text{otherwise}  \end{cases}
\label{eq:wloss2}
\end{equation}

In Eq. \ref{eq:wloss} and Eq. \ref{eq:wloss2}, $\mathcal{R}$ is the set of indices within the pose representation that correspond to rotational components. $P_m$ is referred to as the main time step. For a static pose as input, $P_m$ is the time step at which the pose is inserted into the base motion. In the case of video-extracted motion, $P_m$ is defined by the user and is regarded as the main pose that characterizes the input motion. For instance, if the video-extracted motion is a flying sidekick, $P_m$ should be set to the time step when the person has their kicking leg fully extended and airborne.

The weight $v$ is a positive value greater than one, intended to assign higher importance to $P_m$. $P'$ is a set containing $p$ time steps before and after the insertion of the input motion. Specifically, with video-extracted motion,  $\mathcal{P} = \{P - p, P - p +1, \ldots, P, P + 1, \ldots, P + p,  P + N - p, P + N - p - 1,   \ldots N+P + p \}$. For static pose as input,  $\mathcal{P} = \{P_m - p, P_m - p + 1, \ldots, P_m - 1, P_m + 1, \ldots, P_m + p - 1, P_m + p\}$. The purpose is to provide freedom to the diffusion model to complete the time steps before and after the insertion, ensuring smooth transitions without discontinuities between the base motion and the input motion.

Note that the purpose of predicting $\pmb{x}_0$ in the diffusion model is to enable weighting operations on the motion itself, which is more tractable and intuitive than on the noise or the mean of the noise.

\subsubsection{Diffusion Model Optimization Module}
\label{sec:dmom}

Eq. \ref{eq:loss_nn} defines the loss function used to fine-tune the diffusion model. It is important to note that training over multiple time steps with the combined motion can lead to catastrophic forgetting. That is, where the diffusion model loses its ability to map the base embedding to the base motion. To alleviate this issue, we propose an additional loss term designed to preserve the knowledge of the base motion. This term is weighted by a positive value $\rho$ specified by the user.

\begin{equation}
    L_{f} = \pmb{W}  \odot ||\pmb{x}_0^C  - f_\theta(\pmb{x}_t^{C}, t, \pmb{e}_{opt}) ||^2 + \rho \cdot ||\pmb{x}_0^B  - f_\theta(\pmb{x}_t^{B}, t, \pmb{e}) ||^2
    \label{eq:loss_nn}
\end{equation}

\subsection{Implementation Details}

\subsubsection{Code}
Our code is based on the open-source Guided Diffusion code by OpenAI \cite{guideddiffusion} and uses the pre-trained Motion Diffusion Model \cite{mdm}. It was executed on an Intel i7 11700F at 4.2GHz and an Nvidia RTX 3070 GPU. 

\subsubsection{Hyper-Parameters}

The learning rates were 1e-3 for the first stage and 1e-6 for the second stage. The number of training iterations for both stages -- as well as the probability for training the base motion  -- varied depending on the generated motion, ranging from 100 to 2500 iterations. Similarly, the values for $v$ (Eq. \ref{eq:wloss2}), $\eta$ (Eq. \ref{eq:lce}), and $\rho$ (Eq. \ref{eq:loss_nn}) ranged from 3 to 10, 0 to 1, and 0 to 1, respectively, depending on the generated motion. $\rho$ also varied from 0 to 1. The remaining hyper-parameters -- i.e., the sets $\mathcal{P}$ and $\mathcal{P'}$ and the main time step $P_{m}$ -- also depended on the generated motion.

%\section{Experimental Setup}
%\subfile{sections/setup}

\section{Experiments and Discussion}

The objective of our experiments is to assess the quality of the motions produced by our method. Ideally, these motions should demonstrate the same level of realism as basic actions like walking, running, and jumping -— i.e., motions that are frequently represented in text-motion datasets such as HumanML3D \cite{humanml3d} and thus easier to generate. Additionally, our generated motions should align with the specified poses or video conditions. In the following sections, we outline the experimental setup employed to validate the realism and alignment of our generated motions. The experimental results are then presented and discussed.

\subsection{Setup}

To showcase our method, we opted to generate 16 motions: eight generated from pose insertion (Table \ref{tab:pi}) and eight from video-extracted motion (Table \ref{tab:vi}).

These motions were selected arbitrarily and represent common motions of everyday life that cannot be generated by text-to-motion models due to the limited motion generation capabilities of these models. Additionally, existing motion editing methods (e.g., \cite{mdm, liu2023plan, flame, momask}) cannot generate them, as these edition methods are limited to modifying local characteristics and depend on text prompts for editing. As previously mentioned, this reliance on text prompts significantly restricts the editing capabilities since text-to-motion models have a limited language understanding due to the scarcity of text-motion datasets -- e.g., terms like ``football" or ``carrying a heavy weight" are never seen during training.

\begin{figure*}
    \centering
    \includegraphics[width=0.9\textwidth]{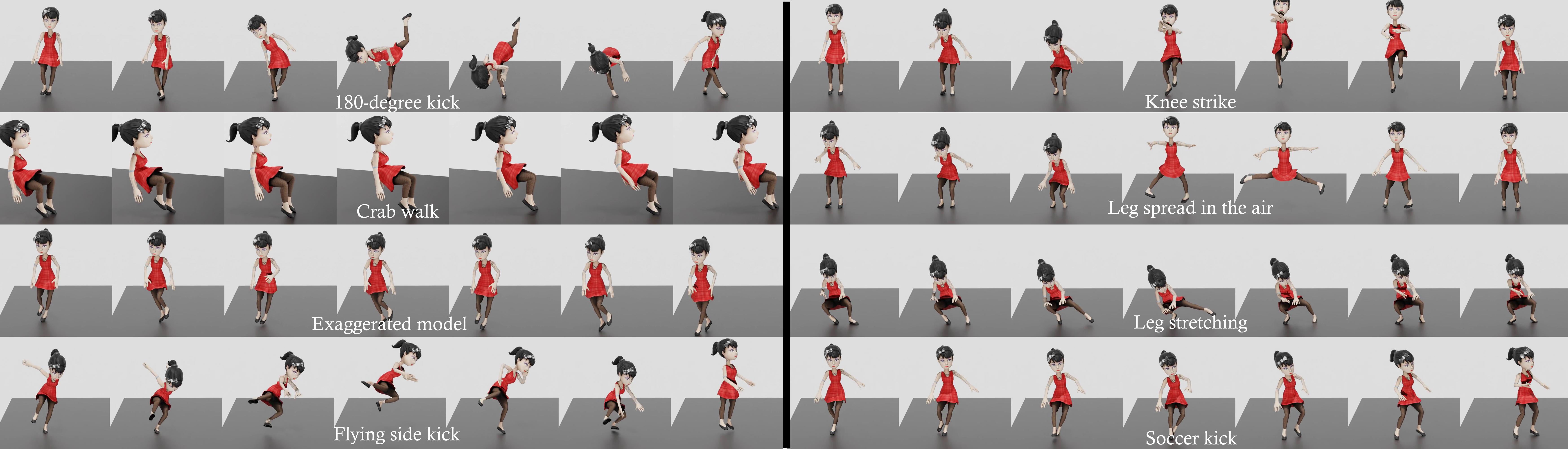}
    \caption{Visualization of some of the motions generated by our method.}
    \label{fig:exp}
\end{figure*}

For comparison, we used MDM \cite{mdm} to generate a set of baseline motions. These motions included walking, running, lying down, sitting down, kicking, performing a cartwheel, squatting, and crawling. Fig. \ref{fig:exp} illustrates a subset of the generated motions with our method.

\begin{table}[tb]
\centering
\small
\setlength{\tabcolsep}{2pt}
\begin{tabular}{llcc}
\textbf{Motion} & \multicolumn{1}{c}{\textbf{Description}}                                                                & \textbf{S} & \textbf{BM} \\ \hline
180$^o$ kick    & \begin{tabular}[c]{@{}l@{}}A high martial arts kick with legs\\ forming a 180-degree angle\end{tabular} & L          & K           \\ \hline
Heavy load     & Walk carrying a heavy and large load     & G & W  \\ \hline
Frog jump      & Jump like a frog                         & G & J  \\ \hline
Jump and knee  & Jump while kneeing an imaginary target   & G & J  \\ \hline
Leg spread     & Jump and spread the legs sideways in air & L & J  \\ \hline
Knee strike    & Jump and strike with the knee            & L & J  \\ \hline
Leg stretching & Squat and stretch the left leg           & L & SQ \\ \hline
One-leg jump    & \begin{tabular}[c]{@{}l@{}}Jump on one leg while keeping the other \\ leg extended behind\end{tabular}  & G          & J           \\ \hline
\end{tabular}
\caption{Details regarding the motions generated by pose insertion. S - Scenario. G - Global. L - Local. BM - Base Motion. W - Walk. J - Jump. SQ - Squat.}
\label{tab:pi}
\end{table}

\begin{table}[]
\centering
\small
\setlength{\tabcolsep}{2pt}
\begin{tabular}{llcc}
\textbf{Motion} & \multicolumn{1}{c}{\textbf{Description}}                                                                   & \textbf{S} & \textbf{BM} \\ \hline
Crab walk       & Move on hands and feet with hips lifted                                                                    & G          & W           \\ \hline
\begin{tabular}[c]{@{}l@{}}Exaggerated\\ model\end{tabular} &
  \begin{tabular}[c]{@{}l@{}}Walk like a model with pronounced \\ hip sways\end{tabular} &
  G &
  W \\ \hline
Flying sidekick & Leap into the air and kick sideways                                                                        & L          & J           \\ \hline
Lunge           & Take a step forward and bend both knees                                                                    & L          & W           \\ \hline
March walk      & \begin{tabular}[c]{@{}l@{}}Walk lifting the knees high and swinging \\ the arms rhythmically\end{tabular}  & G          & W           \\ \hline
Soccer kick     & \begin{tabular}[c]{@{}l@{}}Extend the left arm for balance and kick \\ low with the right leg\end{tabular} & G          & K           \\ \hline
Split leap      & \begin{tabular}[c]{@{}l@{}}Jump and extend one leg forward and\\ the other leg backward\end{tabular}       & L          & J           \\ \hline
\begin{tabular}[c]{@{}l@{}}Toe-touch \\ kick walk\end{tabular} &
  \begin{tabular}[c]{@{}l@{}}Walk, lifting each leg high as if \\ performing a toe-touch kick\end{tabular} &
  G &
  W
\end{tabular}
\caption{Details regarding the motions generated by video insertion. S - Scenario. G - Global. L - Local. BM - Base Motion. W - Walk. J - Jump. K - Kick.}
\label{tab:vi}
\end{table}

The Fréchet Inception Distance (FID) is a widely used metric for assessing the quality of synthetic data. It measures the similarity between the distributions of generated data and real data. However, in our case, since we lack ground truth data corresponding to the motions generated by our method, the FID metric is not applicable. %Moreover, using HumanML3D or KIT data as the real data for FID computation would inevitably produce erroneous results. This is because these datasets do not contain the motions generated by our method, leading to an obvious gap in data distribution.

Following MoMask \cite{momask}, the motions (both generated by our method and the basic motions) were imported into Blender software and reproduced using human-like characters from Mixamo. The videos of these animations were used to visually inspect the quality of the motions. To provide a quantitative evaluation, we conducted an online user study with 26 participants. The participants were shown videos of both the motions generated by our method and the basic motions, without knowing which was which. They were asked to rate the realism of the motions on a scale from 0 to 2, where 0 denoted unrealistic, 1 denoted somewhat realistic, and 2 denoted realistic. Additionally, the participants rated how well the motion aligned with its text description -- i.e., the description of the desired outcome for our  motions or text prompt for the baseline motions. A score of 0 indicated no alignment with the description, 1 indicated partial alignment, and 2 indicated complete alignment. For each motion, the participants also had the option to provide a response indicating why they believed the motion was unrealistic or did not align with the description or text prompt.

\subsection{Results}

\begin{table}[]
\centering
\begin{tabular}{l|cc}
\textbf{Motions}  & \textbf{Realism Score} & \textbf{Alignment Score} \\ \hline
Our  motions & 1.261 $ \pm$ 0.683          & 1.719 $ \pm$ 0.601       \\ \hline
Basic motions  & 1.356 $ \pm$ 0.626          & 1.801 $ \pm$ 0.348  \\   \hline      
\end{tabular}
\caption{Mean $ \pm$ standard deviation of the realism and alignment scores. The maximum score is 2, which represents complete realism or alignment with the text description.}
\label{tab:usr}
\end{table}

Table \ref{tab:usr} details the mean and standard deviation of the realism and alignment scores. Our motions achieve scores similar to the basic motions. %These basic motions are well-represented in the training set used to train the diffusion model, resulting in a model that excels at generating them. Therefore, achieving scores similar to the basic motions is a positive outcome. 

\subsubsection{Realism}

As shown in Fig. \ref{fig:exp}, the motions generated by our method exhibit important characteristics. For instance, in the 180-degree kick, the character bends forward to maintain balance while kicking. Similarly, in the leg stretching motion, the character moves the upper body to stay balanced. In motions involving jumping -- such as the knee strike, legs spread in the air, and flying sidekick -- the character compresses the legs and moves the arms to generate power for the jump.

The open-ended questions provide additional insights into the results. For example, in the flying sidekick motion (Bottom left of Fig. \ref{fig:exp}), one participant noted that one leg did not provide sufficient support for the character, with a more realistic motion being the fall of the character right after the execution of the kick. In the crab walk motion, two participants noted that the arms and hips did not provide adequate support. The character appeared to be moving forward as if sitting on a pretend chair rather than on the ground. A similar concern about falling was also raised in the basic motion of sitting down, where one participant suggested that a more realistic motion would involve the character placing their hands on a surface for support during the sitting.

Three participants also pointed out that certain motions occur too quickly. For instance, in the leg stretching motion, the character extends their leg rather fast. A similar observation was also pointed out in the leg spread motion. However, the concern about the speed of the motion was not restricted to the motions generated by our method since participants also noted that the basic motion of lying down exhibited this issue.

Another aspect of motion realism that participants presented was the discrepancy between upper and lower limb behavior. For example, when walking in an exaggerated model-like manner, one participant observed that the upper body needed more swaying to better match the leg movements. This mismatch concern was also observed in the basic motion of kicking, in which the arms of the character remained stiff during the kicking action.

Overall, in terms of motion realism, our method enables the generation of  motions with similar realism to the basic motions well-represented in the training set. Given a single short video or pose, our method can produce motions without the need for additional data collection and annotation. This implies that our approach is data-efficient. Additionally, it was observed that the concerns related to support, speed, and limb discrepancy are characteristics of the pre-trained diffusion model itself, rather than being specific to our method, since these same concerns were also evident in the basic motions. Note that our method can also be applied with other pre-trained text-to-motion diffusion models, hence it is not restricted to MDM \cite{mdm}. The total execution time of our method varied between two and ten minutes when running on an NVIDIA RTX 3070 GPU.

\subsubsection{Alignment}

The primary issues regarding the alignment between motion and textual description arose in the kicking a football and crab walk motions (Fig. \ref{fig:exp}). For the soccer kick, three participants claimed that the character did not convincingly kick the ball. Descriptions varied, with participants indicating that the motion resembled a side step, a dance move, or evading an invisible object. We attribute this issue to the visualization. Since the motion involves interacting with an object (the ball), it is harder for participants to analyze it when the object is not physically present and the effect of the ball is only imagined. In Fig. \ref{fig:exp}, it is observable that the character performs a swing motion with the right leg -- while raising the left arm for balance -- and uses the instep of the right foot to touch the ball. A side step completes the kicking motion by releasing the remaining momentum.  In the crab walk, one participant observed that the motion resembled someone moving forward while sitting on a chair. This issue stems from the pre-trained diffusion model lacking a robust understanding of physical support, as previously mentioned.

Despite the challenges encountered with these two motions, our generated motions still achieve an average alignment score very close to that of the basic motions. This result indicates that our method successfully generates motions in accordance with the video or pose conditions provided as input by the user. 

%\subsection{Benefits, Limitations, and Future Work}

%Our method allows users to generate motions without resorting to extensive collection and annotation of text-motion data to train a model. Hence, it allows the generation of motions not present in the training set of a text-to-motion model. This is done by editing a base motion according to a single short video clip or pose as condition.

%Unlike other methods \cite{mdm, flame, momask, motionx}, our approach is not restricted to local motion editing (i.e., modifying only local temporal or spatial characteristics of base motion) and supports editing based on video or pose conditions, rather than text. Editing a base motion with a text is limited by the model's language understanding. By using video or pose information as a condition, our method can generate motions not seen in the training set. 

\section{Conclusions}
We proposed a novel approach to alleviate the challenges posed by data scarcity in text-motion datasets, which severely limits the motion diversity achievable by current text-to-motion models. Our method incorporates visual conditions (such as images or videos) alongside textual prompts to  broaden the repertoire of generated motions. Our evaluation demonstrated that the motions produced by our model maintain realism and align well with the specified visual conditions.

As previously discussed, the limitations of our method are a reflection of the pre-trained diffusion model. Future work could involve incorporating an additional hyper-parameter to control motion speed and introducing an extra loss function term to minimize discrepancies in movements between the lower and upper limbs. For additional realism, future work should explore integrating physics principles into the generated motions.

%\subfile{sections/conclusions}

\bibliography{references}

\begin{thebibliography}{21}
\providecommand{\natexlab}[1]{#1}

\bibitem[{Dhariwal and Nichol(2021)}]{guideddiffusion}
Dhariwal, P.; and Nichol, A. 2021.
\newblock Diffusion Models Beat GANs on Image Synthesis.
\newblock \emph{CoRR}, abs/2105.05233.

\bibitem[{Guo et~al.(2023)Guo, Mu, Javed, Wang, and Cheng}]{momask}
Guo, C.; Mu, Y.; Javed, M.~G.; Wang, S.; and Cheng, L. 2023.
\newblock MoMask: Generative Masked Modeling of 3D Human Motions.

\bibitem[{Guo et~al.(2022)Guo, Zou, Zuo, Wang, Ji, Li, and Cheng}]{humanml3d}
Guo, C.; Zou, S.; Zuo, X.; Wang, S.; Ji, W.; Li, X.; and Cheng, L. 2022.
\newblock Generating Diverse and Natural 3D Human Motions From Text.
\newblock In \emph{Proceedings of the IEEE/CVF Conference on Computer Vision and Pattern Recognition (CVPR)}, 5152--5161.

\bibitem[{Ho, Jain, and Abbeel(2020)}]{ddpm}
Ho, J.; Jain, A.; and Abbeel, P. 2020.
\newblock Denoising Diffusion Probabilistic Models.
\newblock \emph{CoRR}, abs/2006.11239.

\bibitem[{Jiang et~al.(2023)Jiang, Chen, Liu, Yu, Yu, and Chen}]{motiongpt}
Jiang, B.; Chen, X.; Liu, W.; Yu, J.; Yu, G.; and Chen, T. 2023.
\newblock MotionGPT: Human Motion as a Foreign Language.
\newblock \emph{arXiv preprint arXiv:2306.14795}.

\bibitem[{Kawar et~al.(2023)Kawar, Zada, Lang, Tov, Chang, Dekel, Mosseri, and Irani}]{imagic}
Kawar, B.; Zada, S.; Lang, O.; Tov, O.; Chang, H.; Dekel, T.; Mosseri, I.; and Irani, M. 2023.
\newblock Imagic: Text-Based Real Image Editing with Diffusion Models.

\bibitem[{Kim, Kim, and Choi(2023)}]{flame}
Kim, J.; Kim, J.; and Choi, S. 2023.
\newblock FLAME: free-form language-based motion synthesis \& editing.
\newblock In \emph{Proceedings of the Thirty-Seventh AAAI Conference on Artificial Intelligence and Thirty-Fifth Conference on Innovative Applications of Artificial Intelligence and Thirteenth Symposium on Educational Advances in Artificial Intelligence}, AAAI'23/IAAI'23/EAAI'23. AAAI Press.
\newblock ISBN 978-1-57735-880-0.

\bibitem[{Kocabas(2021)}]{bodymodelgithub}
Kocabas, M. 2021.
\newblock {G}it{H}ub - mkocabas/body-model-visualizer: {G}{U}{I} for visualization and interactive editing of {S}{M}{P}{L}-family body models ie. {S}{M}{P}{L}, {S}{M}{P}{L}-{X}, {M}{A}{N}{O}, {F}{L}{A}{M}{E}. --- github.com.
\newblock \url{https://github.com/mkocabas/body-model-visualizer}.
\newblock [Accessed 22-05-2024].

\bibitem[{Kocabas et~al.(2021)Kocabas, Huang, Hilliges, and Black}]{pare}
Kocabas, M.; Huang, C.-H.~P.; Hilliges, O.; and Black, M.~J. 2021.
\newblock {PARE}: Part Attention Regressor for {3D} Human Body Estimation.
\newblock In \emph{Proc. International Conference on Computer Vision (ICCV)}, 11127--11137.

\bibitem[{Liu et~al.(2023)Liu, Dai, Wang, Cheng, Tang, and Tong}]{liu2023plan}
Liu, J.; Dai, W.; Wang, C.; Cheng, Y.; Tang, Y.; and Tong, X. 2023.
\newblock Plan, Posture and Go: Towards Open-World Text-to-Motion Generation.
\newblock arXiv:2312.14828.

\bibitem[{Loper et~al.(2023)Loper, Mahmood, Romero, Pons-Moll, and Black}]{smpl}
Loper, M.; Mahmood, N.; Romero, J.; Pons-Moll, G.; and Black, M.~J. 2023.
\newblock \emph{SMPL: A Skinned Multi-Person Linear Model}.
\newblock New York, NY, USA: Association for Computing Machinery, 1 edition.
\newblock ISBN 9798400708978.

\bibitem[{Lugaresi et~al.(2019)Lugaresi, Tang, Nash, McClanahan, Uboweja, Hays, Zhang, Chang, Yong, Lee, Chang, Hua, Georg, and Grundmann}]{mediapipe}
Lugaresi, C.; Tang, J.; Nash, H.; McClanahan, C.; Uboweja, E.; Hays, M.; Zhang, F.; Chang, C.; Yong, M.~G.; Lee, J.; Chang, W.; Hua, W.; Georg, M.; and Grundmann, M. 2019.
\newblock MediaPipe: {A} Framework for Building Perception Pipelines.
\newblock \emph{CoRR}, abs/1906.08172.

\bibitem[{Radford et~al.(2018)Radford, Narasimhan, Salimans, and Sutskever}]{gpt}
Radford, A.; Narasimhan, K.; Salimans, T.; and Sutskever, I. 2018.
\newblock Improving language understanding by generative pre-training.

\bibitem[{Tevet et~al.(2022)Tevet, Gordon, Hertz, Bermano, and Cohen-Or}]{motionclip}
Tevet, G.; Gordon, B.; Hertz, A.; Bermano, A.~H.; and Cohen-Or, D. 2022.
\newblock MotionClip: Exposing human motion generation to clip space.
\newblock In \emph{Computer Vision--ECCV 2022: 17th European Conference, Tel Aviv, Israel, October 23--27, 2022, Proceedings, Part XXII}, 358--374. Springer.

\bibitem[{Tevet et~al.(2023)Tevet, Raab, Gordon, Shafir, Cohen-or, and Bermano}]{mdm}
Tevet, G.; Raab, S.; Gordon, B.; Shafir, Y.; Cohen-or, D.; and Bermano, A.~H. 2023.
\newblock Human Motion Diffusion Model.
\newblock In \emph{The Eleventh International Conference on Learning Representations}.

\bibitem[{van~den Oord, Vinyals, and Kavukcuoglu(2017)}]{vqvae}
van~den Oord, A.; Vinyals, O.; and Kavukcuoglu, K. 2017.
\newblock Neural Discrete Representation Learning.
\newblock \emph{CoRR}, abs/1711.00937.

\bibitem[{Wang et~al.(2023)Wang, Leng, Li, Wu, and Liang}]{fgt2m}
Wang, Y.; Leng, Z.; Li, F. W.~B.; Wu, S.-C.; and Liang, X. 2023.
\newblock Fg-T2M: Fine-Grained Text-Driven Human Motion Generation via Diffusion Model.
\newblock In \emph{Proceedings of the IEEE/CVF International Conference on Computer Vision (ICCV)}, 22035--22044.

\bibitem[{Yang, Chen, and Liao(2023)}]{unipaint}
Yang, S.; Chen, X.; and Liao, J. 2023.
\newblock Uni-Paint: A Unified Framework for Multimodal Image Inpainting with Pretrained Diffusion Model.
\newblock In \emph{Proceedings of the 31st ACM International Conference on Multimedia}, MM '23, 3190–3199. New York, NY, USA: Association for Computing Machinery.
\newblock ISBN 9798400701085.

\bibitem[{Zhang et~al.(2023)Zhang, Zhang, Cun, Huang, Zhang, Zhao, Lu, and Shen}]{t2mgpt}
Zhang, J.; Zhang, Y.; Cun, X.; Huang, S.; Zhang, Y.; Zhao, H.; Lu, H.; and Shen, X. 2023.
\newblock T2M-GPT: Generating Human Motion from Textual Descriptions with Discrete Representations.
\newblock In \emph{Proceedings of the IEEE/CVF Conference on Computer Vision and Pattern Recognition (CVPR)}.

\bibitem[{Zhang et~al.(2022)Zhang, Cai, Pan, Hong, Guo, Yang, and Liu}]{motiondiffuse}
Zhang, M.; Cai, Z.; Pan, L.; Hong, F.; Guo, X.; Yang, L.; and Liu, Z. 2022.
\newblock MotionDiffuse: Text-Driven Human Motion Generation with Diffusion Model.
\newblock arXiv:2208.15001.

\bibitem[{Zhong et~al.(2023)Zhong, Hu, Zhang, and Xia}]{attt2m}
Zhong, C.; Hu, L.; Zhang, Z.; and Xia, S. 2023.
\newblock AttT2M: Text-Driven Human Motion Generation with Multi-Perspective Attention Mechanism.
\newblock In \emph{Proceedings of the IEEE/CVF International Conference on Computer Vision (ICCV)}, 509--519.

\end{thebibliography}

\end{document}